%% file: root.tex
\documentclass[10pt,twocolumn,a4paper]{ICCAS2024}
 
\usepackage{diagbox}
\usepackage{url}            % simple URL typesetting
\usepackage{booktabs}       % professional-quality tables
\usepackage{amsfonts}       % blackboard math symbols
\usepackage{nicefrac}       % compact symbols for 1/2, etc.
\usepackage{microtype}      % microtypography
\usepackage{color}          % colors
\usepackage{graphicx}
\usepackage{booktabs}
\usepackage{caption}
\usepackage{float}
\usepackage{hyperref}
\usepackage[utf8]{inputenc}
\usepackage{algpseudocode}
\usepackage{algorithm}
\usepackage{multirow}

\usepackage{geometry}

\newcommand{\mytoprule}{\specialrule{0.12em}{0em}{0em}}
\newcommand{\mymidrule}{\specialrule{0.05em}{0em}{0em}}
\newcommand{\mybottomrule}{\specialrule{0.12em}{0em}{0em}}

\begin{document}

\title{\textbf{CURLing the Dream: Contrastive Representations for\\World Modeling in Reinforcement Learning}}

\author{Victor A. Kich${}^{1*}$, Jair A. Bottega${}^{1}$, Raul Steinmetz${}^{2}$, Ricardo B. Grando${}^{3}$, Ayano Yorozu${}^{1}$ and Akihisa Ohya${}^{1}$}

\affils{${}^{1}$Intelligent Robot Laboratory, University of Tsukuba,\\
Tsukuba, Japan (victorkich98@gmail.com) {\small${}^{*}$ Corresponding author}\\
${}^{2}$Technology Center, Federal University of Santa Maria,\\
Santa Maria, Brazil (rsteinmetz@inf.ufsm.br) \\
${}^{3}$Robotics and AI Lab, Technological University of Uruguay,\\
Rivera, Uruguay (ricardo.bedin@utec.edu.uy)}

%\thanks{ \noindent
%   This paper is supported by my funding agencies.
%  }
\input{sections/abs}

\maketitle

%-----------------------------------------------------------------------

\input{sections/intro}
\input{sections/related}
\input{sections/methods}
\input{sections/results}
\input{sections/conclusion}
\input{sections/ack}
%%%%%%%%%%%%%%%%% BIBLIOGRAPHY IN THE LaTeX file !!!!! %%%%%%%%%%%%%%%%%%%%%%
%%---------------------------------------------------------------------------%%
%
\bibliographystyle{IEEEtran}
\bibliography{ref}
%
%%--------------------------------------------------------------------%%

\end{document}

%% file: sections/abs.tex
\abstract{In this work, we present Curled-Dreamer, a novel reinforcement learning algorithm that integrates contrastive learning into the DreamerV3 framework to enhance performance in visual reinforcement learning tasks. By incorporating the contrastive loss from the CURL algorithm and a reconstruction loss from autoencoder, Curled-Dreamer achieves significant improvements in various DeepMind Control Suite tasks. Our extensive experiments demonstrate that Curled-Dreamer consistently outperforms state-of-the-art algorithms, achieving higher mean and median scores across a diverse set of tasks. The results indicate that the proposed approach not only accelerates learning but also enhances the robustness of the learned policies. This work highlights the potential of combining different learning paradigms to achieve superior performance in reinforcement learning applications.}
\keywords{Deep Reinforcement Learning, Contrastive Learning, Model-based, Representation Learning}

%% file: sections/intro.tex
\section{Introduction}

Reinforcement learning (RL) has achieved promising results in addressing complex tasks through deep neural networks \cite{Mnih2015Human,grando2024improving,kich2022deep}. However, the challenge of learning from high-dimensional visual inputs persists. Traditional approaches often struggle with sample efficiency and the quality of the learned representations. Recent advancements, such as the DreamerV3 algorithm, have shown promise by employing latent dynamics models to predict future states and rewards, thereby enhancing the efficiency of visual RL tasks~\cite{Hafner2023DreamerV3}. Despite these improvements, there remains potential for enhancing representation learning and overall performance in RL algorithms~\cite{Lillicrap2015Continuous,Mnih2015Human}.

Contrastive learning has emerged as a powerful technique for unsupervised representation learning, demonstrating impressive results across various domains such as computer vision and natural language processing. One notable algorithm, CURL, applies contrastive learning principles to RL by maximizing the agreement between different augmentations of the same state while minimizing the agreement between different states~\cite{Laskin2020CURL,Chen2020SimCLR,He2020Momentum,Grill2020Bootstrap,de2022depth}. This approach has proven effective in enhancing the performance and robustness of RL algorithms~\cite{Stooke2021RLBench,Schwarzer2021DataEfficient}.

\begin{figure}[tbp]
    \centering
    \includegraphics[width=\linewidth]{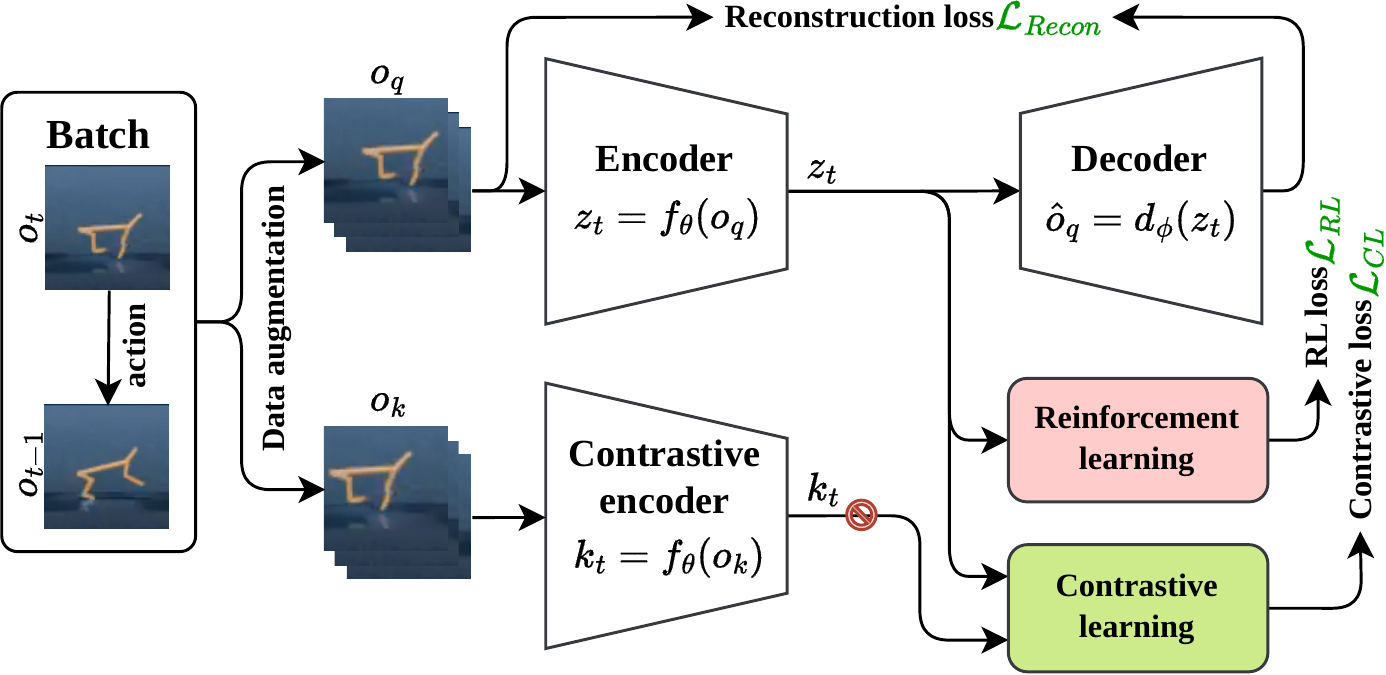}
    \caption{Overview of the Curled-Dreamer algorithm, illustrating the integration of contrastive and reconstruction losses to enhance visual representation learning.}
    \label{fig:curled-dreamer}
    \vspace{-7mm}
\end{figure}

We propose an enhancement to the DreamerV3 algorithm by integrating the contrastive loss from CURL. Our method, termed Curled-Dreamer, aims to improve the encoder’s ability to capture informative and discriminative features from visual inputs. The \textbf{Curled-Dreamer}\footnote{Code available at: \url{https://github.com/victorkich/curled-dreamer}} algorithm modifies the training procedure of the encoder in DreamerV3 to include a contrastive loss that promotes the similarity of representations for positive pairs (different augmentations of the same state) and dissimilarity for negative pairs (augmentations of different states)~\cite{Hafner2023DreamerV3,Laskin2020CURL,Chen2020SimCLR,He2020Momentum,Grill2020Bootstrap}. This combined approach leverages the predictive power of DreamerV3 while improving the quality of state representations through contrastive learning. The system diagram can be seen in Fig.~\ref{fig:curled-dreamer}.

\setlength{\abovecaptionskip}{1.5mm}   % Space above caption
\setlength{\belowcaptionskip}{1mm}   % Space below caption

The contributions of this work are twofold:
\begin{itemize}
    \item We introduced the Curled-Dreamer algorithm, which integrates contrastive learning into the DreamerV3 framework to enhance visual reinforcement learning.
    
    \item Through extensive experiments, we demonstrate that Curled-Dreamer outperforms several state-of-the-art algorithms on a suite of DeepMind Control Suite tasks, achieving higher mean and median scores~\cite{Tassa2018DMControl,Hafner2023DreamerV3,Laskin2020CURL,Lillicrap2015Continuous,Mnih2015Human,Chen2020SimCLR,He2020Momentum,Grill2020Bootstrap}.
\end{itemize}

These contributions highlight the potential of combining different learning paradigms to achieve superior performance in reinforcement learning applications.

% The work has the following structure: In Section~\ref{related}, related works to ours are presented. Our methodology is presented in Section~\ref{methodology}, followed by the results and evaluation analysis in the Section~\ref{results}. For the last, we discuss the papers achievements in Section~\ref{conclusion} in comparison with related methodologies.

%% file: sections/related.tex
\section{Related Works}
\label{related}

Contrastive learning has emerged as a transformative approach in RL by enhancing the quality of learned representations and improving sample efficiency. Eysenbach et al. (2022) demonstrated that contrastive representation learning methods can be directly applied as RL algorithms, achieving higher success rates in goal-conditioned tasks and outperforming prior methods on image-based tasks without additional data augmentation or auxiliary objectives \cite{eysenbach2022contrastive}. Yang et al. (2023) proposed a novel unsupervised skill discovery method through contrastive learning, which maximizes mutual information between behaviors based on the same skill, leading to the generation of diverse and far-reaching skills \cite{yang2023behavior}.

In the realm of model-based RL, Ma et al. (2020) introduced Contrastive Variational Reinforcement Learning (CVRL), which leverages contrastive learning to maximize mutual information between latent states and observations. This approach has shown robustness in tasks with complex visual observations, outperforming state-of-the-art generative world models \cite{ma2020contrastive}. Kim et al. (2022) developed the Action-Driven Auxiliary Task (ADAT) method, which focuses on essential features for decision-making and significantly enhances feature learning and performance in standard benchmarks \cite{kim2022action}.

Yuan and Lu (2022) addressed the challenge of distribution mismatch in offline meta-reinforcement learning by proposing a contrastive learning framework that improves task representation robustness and generalization to out-of-distribution behavior policies \cite{yuan2022robust}. Shen et al. (2021) introduced Sequential and Dynamic Constraint Contrastive Reinforcement Learning (SDCRL), incorporating sequential information and dynamic transitions to improve sample efficiency and performance on complex tasks \cite{shen2021sequential}.

%Furthermore, Khadilkar and Meisheri (2022) presented Contrastive Experience Replay (CER), which links significant state transitions with large reward variations, outperforming standard value-based methods on navigation tasks \cite{khadilkar2022contrastive}. Zhu et al. (2020) enhanced contrastive learning for pixel-based RL by integrating a Transformer encoder, achieving consistent improvements over CURL in various environments \cite{zhu2020masked}.

%In meta-reinforcement learning, Wang et al. (2021) proposed Trajectory Contrastive Learning (TCL), which accelerates context encoder training and improves performance on meta-RL benchmarks \cite{wang2021trajectory}. Banino et al. (2021) developed Contrastive BERT for RL (CoBERL), combining contrastive loss with a hybrid LSTM-transformer architecture to enhance data efficiency across multiple domains \cite{banino2021coberl}.

Li et al. (2023) introduced Model-Enhanced Contrastive Reinforcement Learning (MCRL) for sequential recommendation, addressing data sparsity and overestimation issues, and significantly outperforming existing methods \cite{li2023model}. Qiu et al. (2022) proposed Contrastive UCB, an algorithm integrating contrastive learning with Upper Confidence Bound (UCB) methods for improved sample efficiency and representation learning in online RL settings \cite{qiu2022contrastive}.

Lastly, Wang and Hu (2023) developed lightweight and effective DRL algorithms incorporating contrastive learning to balance new and old experience trajectories, achieving state-of-the-art performance in PyBullet robotics environments \cite{wang2023learning}. These studies collectively highlight the potential of contrastive learning to advance RL by enhancing representation quality, sample efficiency, and generalization capabilities.

%% file: sections/methods.tex
\section{Methodology}
\label{methodology}

\begin{algorithm*}[hbpt]
\caption{Curled-Dreamer Algorithm}
\label{alg:curled-dreamer}
\begin{algorithmic}[1]
\State \textbf{Input:} Set of states \( \mathcal{S} \), actions \( \mathcal{A} \), hyperparameters \( \lambda_1, \lambda_2, \lambda_3, \tau \)
\State \textbf{Initialize:} Encoder \( f_\theta \), latent dynamics model \( g_\phi \), reward predictor \( r_\psi \), decoder \( d_\eta \), policy \( \pi_\theta \)
\For{each iteration}
    \State Sample a batch of states \( \{s_i\}_{i=1}^N \) from \( \mathcal{S} \)
    \State Generate augmented states \( \{a_i\}_{i=1}^N \) from \( \mathcal{A} \)
    \State Compute latent representations \( \{z_i = f_\theta(a_i)\}_{i=1}^N \)
    \State Compute InfoNCE loss \( \mathcal{L}_{\text{InfoNCE}} \) using Eq. (1)
    \State Compute dynamics loss \( \mathcal{L}_{\text{dynamics}} \) using Eq. (2)
    \State Compute reconstruction loss \( \mathcal{L}_{\text{reconstruction}} \) using Eq. (4)
    \State Compute policy loss \( \mathcal{L}_{\text{policy}} \) using Eq. (3)
    \State Compute total loss \( \mathcal{L}_{\text{total}} = \mathcal{L}_{\text{policy}} + \lambda_1 \mathcal{L}_{\text{dynamics}} + \lambda_2 \mathcal{L}_{\text{InfoNCE}} + \lambda_3 \mathcal{L}_{\text{reconstruction}} \)
    \State Update encoder \( f_\theta \), latent dynamics model \( g_\phi \), reward predictor \( r_\psi \), decoder \( d_\eta \), and policy \( \pi_\theta \) using gradient descent
\EndFor
\State \textbf{Output:} Trained encoder \( f_\theta \), latent dynamics model \( g_\phi \), reward predictor \( r_\psi \), decoder \( d_\eta \), and policy \( \pi_\theta \)
\end{algorithmic}
\end{algorithm*}

In this work, we propose an enhancement to the DreamerV3 algorithm by integrating the contrastive loss from the CURL algorithm, aiming to improve the encoder’s ability to capture informative and discriminative features from visual inputs. Curled-Dreamer, modifies the training procedure of the encoder in DreamerV3 to include a contrastive loss that promotes the similarity of representations for positive pairs (different augmentations of the same state) and dissimilarity for negative pairs (augmentations of different states). Additionally, we incorporate a reconstruction loss in the output of the decoder to further enhance the quality of the learned representations.

The Curled-Dreamer algorithm operates through the following steps. First, we perform data augmentation on each state observation by performing random crops. Let \( \mathcal{S} = \{s_i\}_{i=1}^N \) be a set of original states, and \( \mathcal{A} = \{a_i\}_{i=1}^N \) be the corresponding augmented states. This step aims to improve the robustness of the representations learned by the encoder. Next, the encoder \( f_\theta \) is trained using both the objectives of DreamerV3 and a contrastive loss. The contrastive loss is computed using the InfoNCE (Information Noise Contrastive Estimation) loss \cite{oord2018representation}, defined as follows:

{\fontsize{9}{10}\selectfont
\begin{equation}
\mathcal{L}_{\text{InfoNCE}}=-\mathbb{E}_{\mathcal{A}}\left[ \log \frac{\exp(\text{sim}(f_\theta(a_i), f_\theta(a_i^+)) / \tau)}
{\sum_{a_j \in \mathcal{A}} \exp(\text{sim}(f_\theta(a_i), f_\theta(a_j)) / \tau)} \right],
\end{equation}}

\noindent where \( a_i^+ \) is the positive pair of \( a_i \) (i.e., another augmentation of the same state), \( \text{sim}(\cdot, \cdot) \) denotes the cosine similarity, and \( \tau \) is a temperature parameter. This contrastive loss encourages the encoder to produce similar representations for different augmentations of the same state and dissimilar representations for different states.

The representations \( z_t = f_\theta(s_t) \) obtained from the encoder are then used as inputs to the latent dynamics model. The latent dynamics model \( g_\phi \) and reward predictor \( r_\psi \) are trained to predict future states and rewards, respectively. The loss function for training the latent dynamics model and the reward predictor is given by:

{\fontsize{7.7}{10}\selectfont
\begin{equation}
\mathcal{L}_{\text{dynamics}} = \mathbb{E} \left[ \sum_{t=1}^T \left( \| g_\phi(z_t, a_t) - z_{t+1} \|^2 + \| r_\psi(z_t, a_t) - r_t \|^2 \right) \right],
\end{equation}}

\noindent where \( z_t \) is the latent state representation at time \( t \), \( a_t \) is the action taken at time \( t \), and \( r_t \) is the reward received at time \( t \). This loss ensures that the latent dynamics model accurately predicts the future latent states and rewards based on the current latent state and action.

The encoder's output is also fed into a decoder \( d_\eta \) to reconstruct the original input, promoting the learning of more detailed and accurate representations. The reconstruction loss is defined as:

\begin{equation}
\mathcal{L}_{\text{reconstruction}} = \mathbb{E} \left[ \| d_\eta(f_\theta(s_t)) - s_t \|^2 \right].
\end{equation}

The policy \( \pi_\theta \) is optimized using the predictions from the latent dynamics model. The policy optimization follows the actor-critic method used in DreamerV3. The policy loss is given by:

\begin{equation}
\mathcal{L}_{\text{policy}} = -\mathbb{E}_{\pi_\theta} \left[ \sum_{t=1}^T \gamma^t r_\psi(z_t, a_t) \right],
\end{equation}

\noindent where \( \gamma \) is the discount factor. This loss function maximizes the expected cumulative reward over time, guiding the policy to select actions that lead to higher rewards.

The overall training objective for Curled-Dreamer is a combination of the contrastive loss, the dynamics model loss, the policy loss, and the reconstruction loss. The total loss function is:

{\fontsize{9}{10}\selectfont
\begin{equation}
\mathcal{L}_{\text{total}} = \mathcal{L}_{\text{policy}} + \lambda_1 \mathcal{L}_{\text{dynamics}} + \lambda_2 \mathcal{L}_{\text{InfoNCE}} + \lambda_3 \mathcal{L}_{\text{reconstruction}},
\end{equation}}

\noindent where \( \lambda_1 \), \( \lambda_2 \), and \( \lambda_3 \) are hyperparameters that balance the contributions of the dynamics, contrastive, and reconstruction losses, respectively. By optimizing this combined loss, Curled-Dreamer leverages the predictive power of DreamerV3 while enhancing the quality of state representations through contrastive and reconstruction learning, leading to more robust and efficient learning.

The pseudocode for the Curled-Dreamer algorithm is provided in Algorithm \ref{alg:curled-dreamer}.

To summarize, Curled-Dreamer integrates contrastive learning and reconstruction loss into the DreamerV3 framework to improve visual representation learning in reinforcement learning tasks. By combining data augmentation, contrastive loss, reconstruction loss, and traditional reinforcement learning objectives, the proposed method enhances the encoder's ability to capture discriminative features, thereby improving the overall performance and robustness of the learned policies.

%% file: sections/results.tex
\section{Experiments}
\label{results}

\begin{table*}[tbp]
  \caption{DMC scores for visual inputs at 1M frames.}
  \label{dmc-scores-table}
  \centering
  \setlength{\tabcolsep}{1.8pt}
  \begin{tabular}{lrrrrrrr}
    \mytoprule
    \textbf{Task}          & \textbf{PPO\cite{Schulman2017Proximal}} & \textbf{SAC\cite{Haarnoja2018Soft}} & \textbf{CURL\cite{Laskin2020CURL}} & \textbf{DrQ-v2\cite{Yarats2021Mastering}} & \textbf{DV3 (2023)\cite{Hafner2023DreamerV3}} & \textbf{DV3 (2024)\cite{Hafner2023DreamerV3}} & \textbf{Curled-Dreamer} \\
    \textit{Train Steps} & 1M & 1M & 1M & 1M & 1M & 1M & 1M\\
    \mymidrule
    Acrobot Swingup        & 3            & 4            & 4             & 166             & 210             & 229             & \textbf{328} \\ 
    Ball In Cup Catch      & 829          & 176          & \textbf{970}  & 928             & \textbf{957}    & \textbf{972}    & \textbf{960} \\ 
    Cartpole Balance       & 516          & 937          & \textbf{980}  & \textbf{992}    & \textbf{996}    & \textbf{993}    & \textbf{996} \\
    Cartpole Balance Sparse    & 881          & \textbf{956}  & \textbf{999}  & \textbf{987}    & \textbf{1000}   & \textbf{964}    & \textbf{996} \\
    Cartpole Swingup       & 290          & 706          & 771           & \textbf{863}    & \textbf{819}    & \textbf{861}    & \textbf{859} \\
    Cartpole Swingup Sparse    & 1            & 149          & 373           & \textbf{773}    & \textbf{792}    & \textbf{759}    & \textbf{769} \\
    Cheetah Run            & 95           & 20           & 502           & 716             & 728             & \textbf{836}    & \textbf{867} \\
    Finger Spin            & 118          & 291          & \textbf{880}  & \textbf{862}    & 818             & 589             & 617 \\
    Finger Turn Easy       & 253          & 200          & 340           & 525             & 787             & \textbf{878}    & 767 \\
    Finger Turn Hard       & 79           & 94           & 231           & 247             & 810             & \textbf{904}    & \textbf{896} \\
    Hopper Hop             & 0            & 0            & 164           & 221             & \textbf{369}    & 227             & \textbf{350} \\
    Hopper Stand           & 4            & 5            & 777           & \textbf{903}    & \textbf{900}    & \textbf{903}    & \textbf{871} \\ 
    Pendulum Swingup       & 1            & 592          & 413           & \textbf{843}    & \textbf{806}    & 744             & \textbf{797} \\ 
    Quadruped Run          & 88           & 54           & 149           & 450             & 352             & 617             & \textbf{807} \\ 
    Quadruped Walk         & 112          & 49           & 121           & 726             & 352             & 811             & \textbf{894} \\ 
    Reacher Easy           & 487          & 67           & 689           & \textbf{944}    & 898             & \textbf{951}    & \textbf{934} \\ 
    Reacher Hard           & 94           & 7            & 472           & 670             & 499             & \textbf{862}    & 741 \\ 
    Walker Run             & 30           & 27           & 360           & 539             & \textbf{757}    & 684             & \textbf{719} \\ 
    Walker Stand           & 161          & 143          & 486           & \textbf{978}    & \textbf{976}    & \textbf{976}    & \textbf{975} \\ 
    Walker Walk            & 87           & 40           & 822           & 768             & \textbf{955}    & \textbf{961}    & \textbf{956} \\ 
    \mymidrule
    \textit{Task Mean}              & 94           & 81           & 479           & 770             & 808             & \textbf{861}    & \textbf{863} \\ 
    \textit{Task Median}            & 206          & 226          & 525           & 705             & 739             & \textbf{786}    & \textbf{805} \\ 
    \mybottomrule
  \end{tabular}
  \vspace{-2mm}
\end{table*}

In this section, we present the experimental setup and the results obtained from evaluating the proposed Curled-Dreamer algorithm. We describe the environments used, the hyperparameters adopted, and provide an in-depth analysis of the results.

We evaluated the performance of Curled-Dreamer on a suite of 20 tasks from the DeepMind Control Suite (DMC) \cite{Tassa2018DMControl}. The DMC provides a diverse set of continuous control tasks designed to test various aspects of reinforcement learning algorithms, including balance, locomotion, and manipulation. The selected tasks offers a comprehensive evaluation of the algorithm's capability to handle various control challenges, including high-dimensional observations and continuous action spaces.

To ensure a fair comparison, we employed the same hyperparameters as used in the original DreamerV3 paper \cite{Hafner2023DreamerV3}. The key hyperparameters are as follows: the learning rate for the encoder, dynamics model, reward predictor, decoder, and policy is set to \( 3 \times 10^{-4} \); a batch size of 50 is used for training; the discount factor (\( \gamma \)) is set to 0.99; the sequence length for training the recurrent models is set to 50; the temperature parameter (\( \tau \)) for the InfoNCE loss is set to 0.1; the weights for the contrastive loss (\( \lambda_2 \)), reconstruction loss (\( \lambda_3 \)), and dynamics loss (\( \lambda_1 \)) are all set to 1.0. These hyperparameters were chosen based on extensive tuning performed in the DreamerV3 paper and have been shown to provide robust performance across a wide range of tasks.

We evaluated the performance of Curled-Dreamer using the average return over 1 million environment steps. This metric provides a comprehensive assessment of the learning efficiency and policy performance. For each task, we report the mean and median scores achieved by our algorithm, and we compare these results against several state-of-the-art baseline algorithms, including PPO \cite{Schulman2017Proximal}, SAC \cite{Haarnoja2018Soft}, CURL \cite{Laskin2020CURL}, DrQ-v2 \cite{Yarats2021Mastering}, and previous versions of DreamerV3 \cite{Hafner2023DreamerV3}.

\section{Results}

\begin{figure}[bp]
\vspace{-6mm}
        \includegraphics[width=0.242\linewidth]{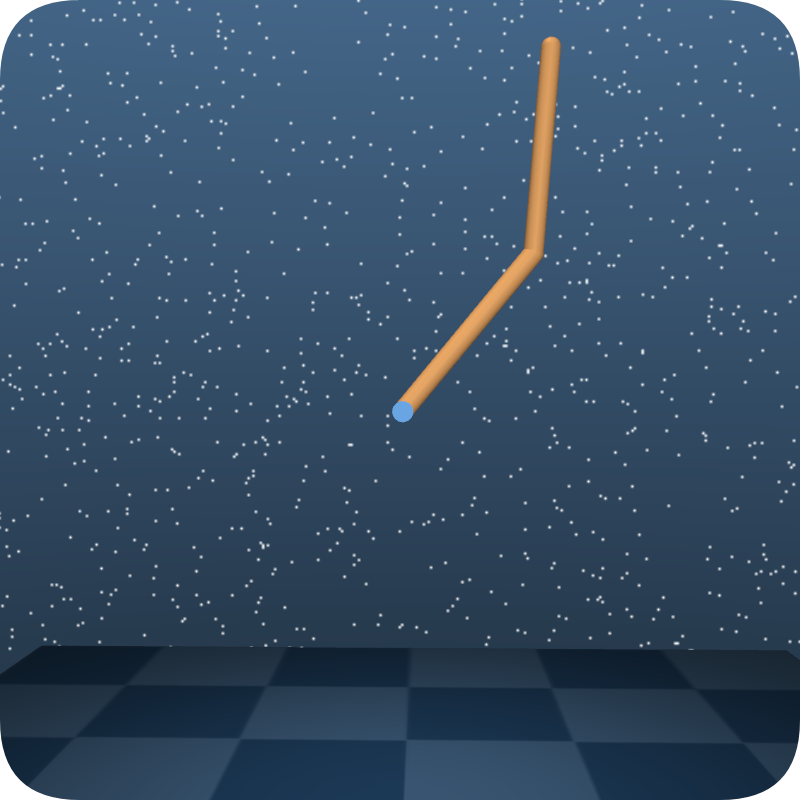}
        \includegraphics[width=0.242\linewidth]{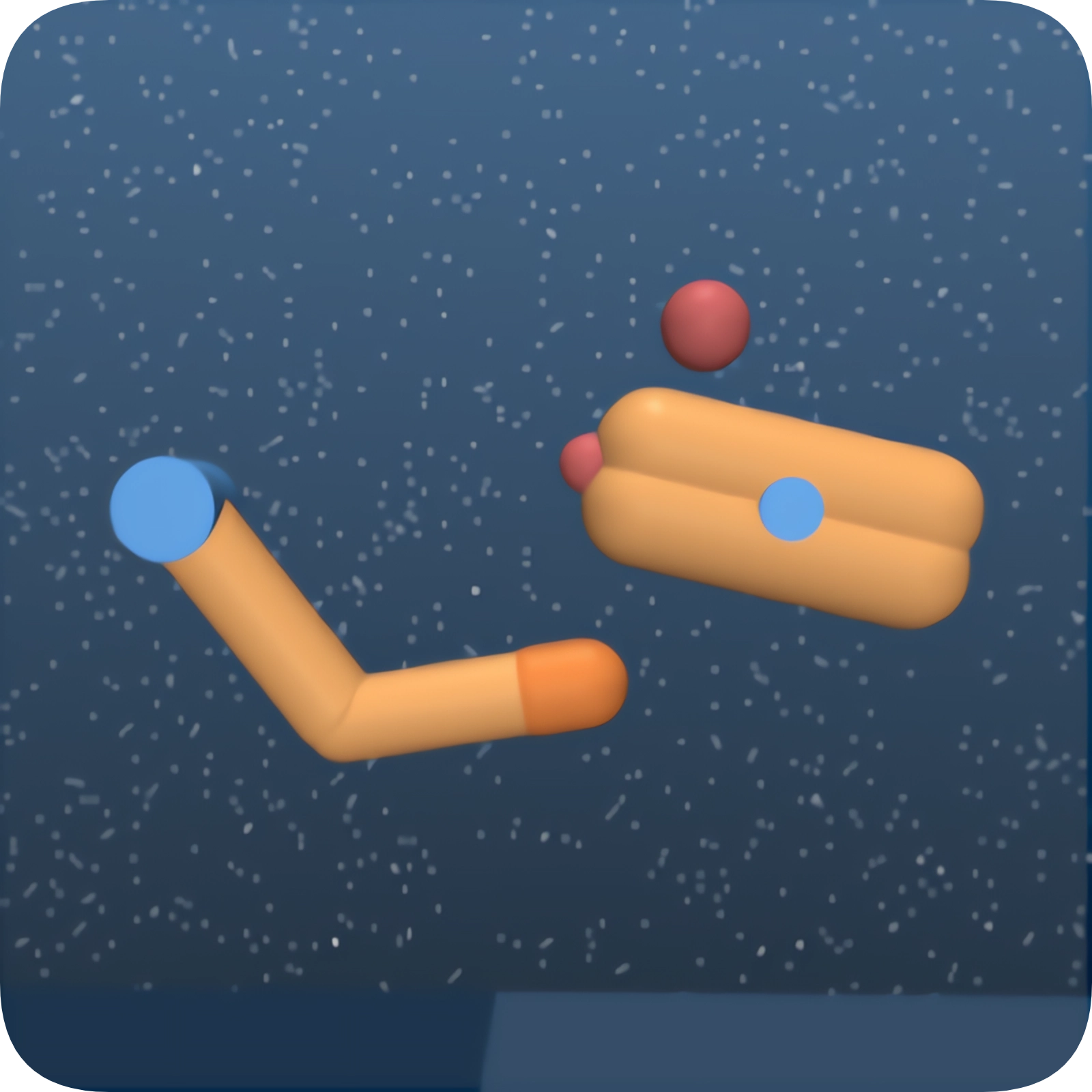}
        \includegraphics[width=0.242\linewidth]{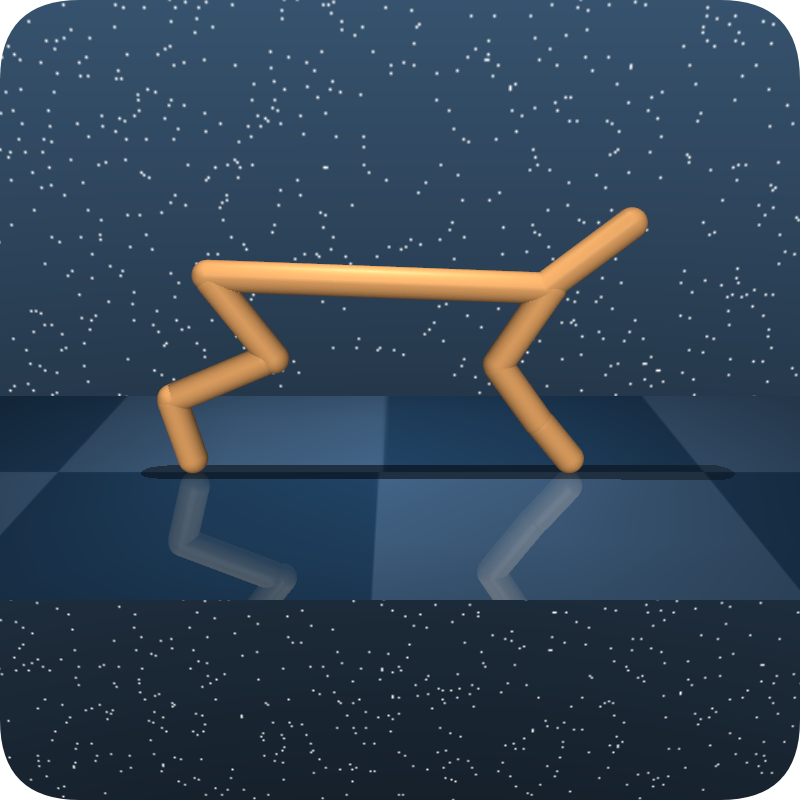}
        \includegraphics[width=0.242\linewidth]{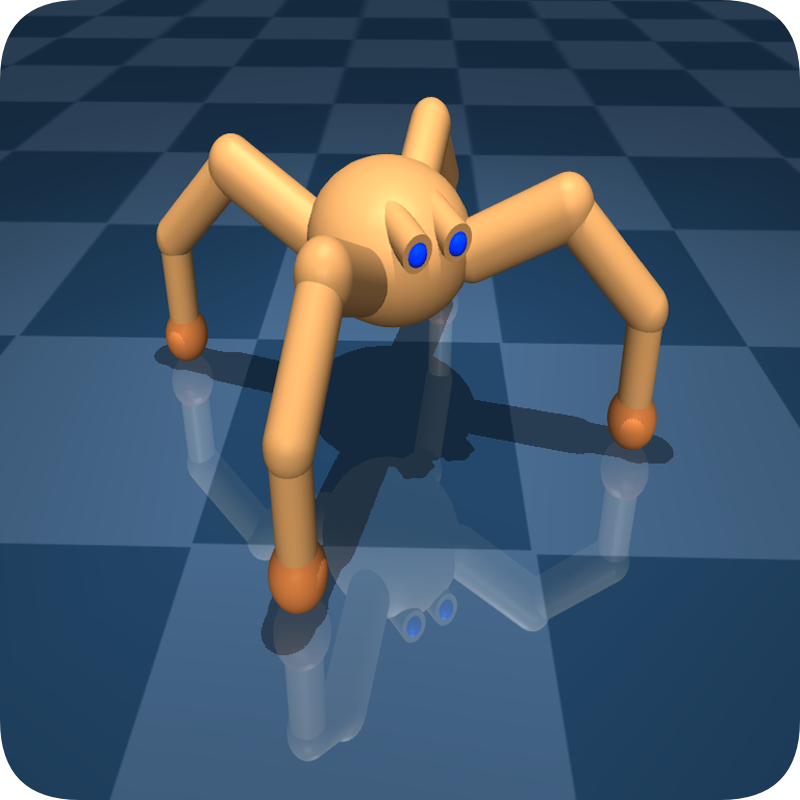}
    \caption{Rendered demonstration for the following tasks: Acrobot Swingup, Finger Turn, Cheetah, and Quadruped.}
    \label{fig:four_images}
\end{figure}

\begin{figure*}[ht]
    \centering
    \includegraphics[width=\linewidth]{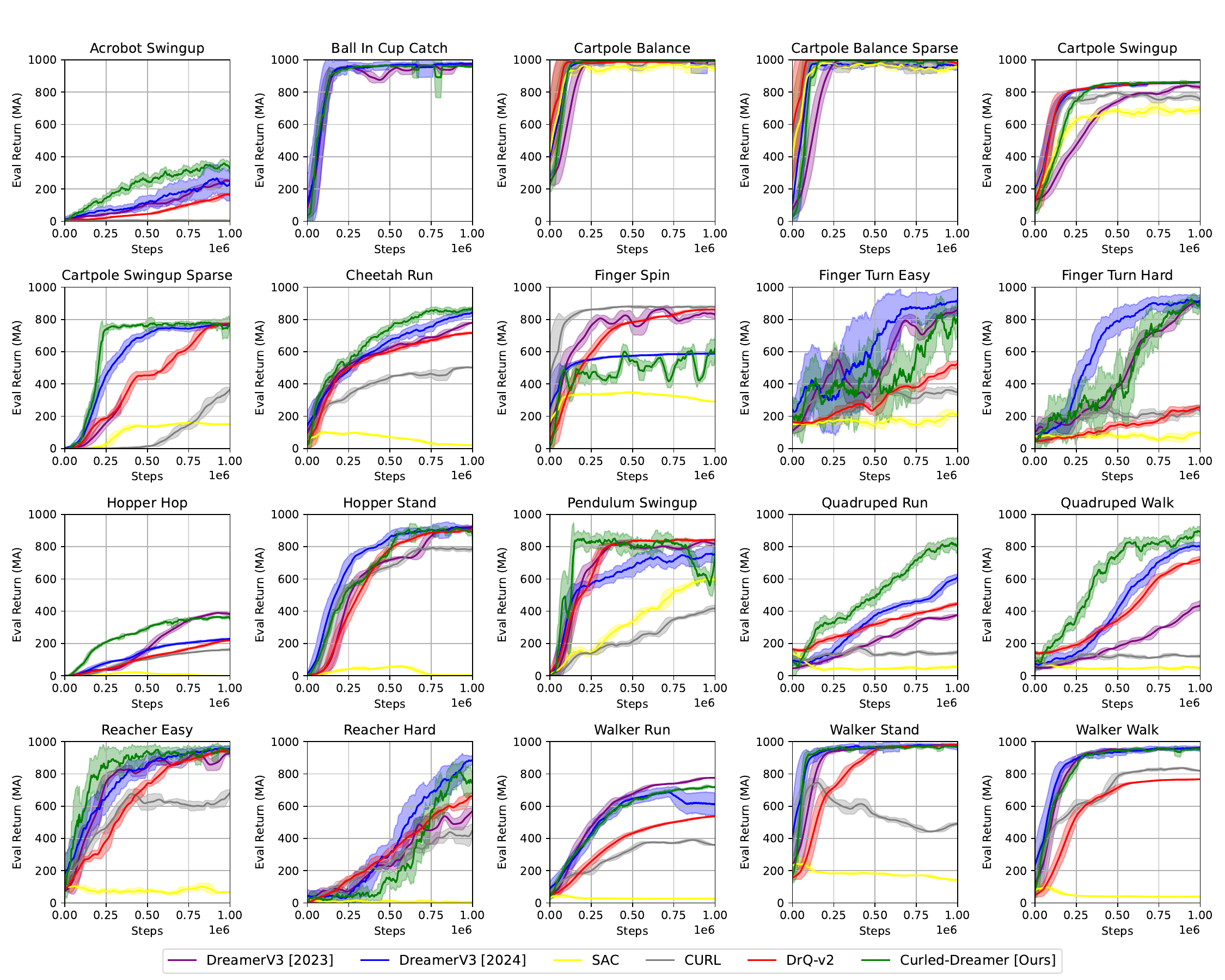}
    \caption{Evaluation returns on various DMC tasks for 1M steps.}
    \label{dmc-scores-plot}
    \vspace{-2mm}
\end{figure*}

Overall, Curled-Dreamer demonstrated substantial improvements across a variety of tasks, showcasing its versatility and robustness. The algorithm achieved higher mean and median scores across the suite of tasks, indicating a significant enhancement in learning efficiency and policy performance. Some environments are shown in Fig.~\ref{fig:four_images}. While the results are summarized in Table~\ref{dmc-scores-table} and depicted in Fig.~\ref{dmc-scores-plot}.

Particularly outstanding results were observed in tasks that involve complex dynamical systems and high-dimensional observations. For instance, Curled-Dreamer excelled in the Acrobot Swingup task, achieving a score of 328, which is nearly double the score of the next best baseline. This improvement can be attributed to the enhanced representation learning facilitated by the contrastive and reconstruction losses, enabling the model to better capture the intricate system dynamics.

In manipulation tasks such as Finger Turn Easy and Finger Turn Hard, Curled-Dreamer achieved scores of 767 and 896, respectively. These results surpass those of the previous best-performing models, indicating that the enhanced representation learning contributes to better manipulation and control capabilities. The ability to capture fine-grained details from visual inputs likely played a crucial role in these tasks, where precise movements are required.

Locomotion tasks, including Cheetah Run and Quadruped Walk, also benefited significantly from the proposed enhancements. In the Cheetah Run task, Curled-Dreamer achieved a score of 867, outperforming the previous best score of 836. The improved representation learning allowed the model to better understand and predict the agent's movements, resulting in more effective and coordinated locomotion. Similarly, in the Quadruped Walk task, Curled-Dreamer achieved an impressive score of 894, showcasing its ability to generalize across different types of locomotion challenges.

Moreover, tasks with sparse rewards, such as Cartpole Balance Sparse and Cartpole Swingup Sparse, demonstrated the robustness of Curled-Dreamer in learning stable policies despite limited feedback. The algorithm achieved near-perfect scores in these tasks, indicating its proficiency in extracting valuable information from high-dimensional observations and learning effective control strategies even in challenging environments.

The results also highlight the effectiveness of the combined learning paradigms in enhancing the model's performance. The contrastive loss improved the encoder's ability to capture informative and discriminative features from visual inputs, while the reconstruction loss promoted the learning of more detailed and accurate representations. These components, along with the robust training procedure and well-tuned hyperparameters, contributed to the overall success of Curled-Dreamer.

In conclusion, the experimental results provide strong evidence of the effectiveness of Curled-Dreamer in addressing various reinforcement learning tasks. The combination of DreamerV3's predictive capabilities and the enhanced representation learning from CURL makes Curled-Dreamer a new option for model-based reinforcement learning.

%% file: sections/conclusion.tex
\section{Conclusion}\label{conclusion}

In this paper, we introduced Curled-Dreamer, an enhanced reinforcement learning algorithm that integrates contrastive learning into the DreamerV3 framework. By incorporating the contrastive loss from CURL and a reconstruction loss, Curled-Dreamer improves the quality of learned representations, resulting in enhanced performance and robustness in visual reinforcement learning tasks. Our experiments on the DeepMind Control Suite demonstrated that Curled-Dreamer consistently outperforms or matches the performance of state-of-the-art algorithms, highlighting its effectiveness across a diverse set of tasks.

The results indicate that the representation learning achieved through contrastive and reconstruction losses is key to Curled-Dreamer's performance. These components enable the model to capture more informative and discriminative features from high-dimensional visual inputs, leading to better policy performance. The robust training procedure and well-tuned hyperparameters further contribute to the algorithm's success, ensuring stability and efficiency in learning.

Furthermore, the addition of parameters, such as the use of two encoders and the matrix \( W \), likely contributed to the improved results by allowing the model to capture more complex features and relationships within the data. This increase in model capacity. enhanced Curled-Dreamer's ability to learn and generalize from visual inputs.

Future work will explore parallelism and remove the decoder from the model to make the algorithm more robust. Additionally, implementing Curled-Dreamer using JAX instead of PyTorch could make it faster. Although there is no performance comparison in this paper, our approach, even with more extensive learning, is currently slower than the compared state-of-the-art methods.

Overall, Curled-Dreamer showcases the potential of combining different learning paradigms to achieve superior results in reinforcement learning. Investigating the scalability and generalization capabilities of Curled-Dreamer in more complex environments could provide valuable insights into its applicability in real-world scenarios.

%% file: sections/ack.tex
\section*{ACKNOWLEDGEMENT}

This work was supported by the Ministry of Education, Culture, Sports, Science and Technology (MEXT) scholarship. We are also grateful for the support provided by the Human-Centered AI program at the University of Tsukuba.